\pdfoutput=1

\documentclass[11pt]{article}

\usepackage[final]{acl}

\usepackage[T1]{fontenc}
\usepackage{times}
\usepackage{latexsym}
\usepackage{csquotes}
\usepackage{booktabs}
\usepackage{multicol}
\usepackage{multirow}
\usepackage{graphicx}
\usepackage{subcaption}
\usepackage{amsmath}
\usepackage{cleveref}
\usepackage{tabularx}
\usepackage{amssymb}
\usepackage[utf8]{inputenc}
\usepackage[ruled,vlined]{algorithm2e}

\usepackage{listings,xcolor}
\usepackage[most]{tcolorbox}

\newtcblisting[auto counter]{codelisting}[2][]{sharp corners,
    fonttitle=\bfseries, colframe=gray, listing only,
    listing options={basicstyle=\small\ttfamily, breaklines=true, breakautoindent=false, breakindent=0pt, escapeinside={(*@}{@*)}
},
    title=Prompt segment \thetcbcounter: #2, #1}
    
\usepackage{microtype}

\usepackage{inconsolata}

\usepackage{graphicx}

\title{Tracr-Injection: Distilling Algorithms into Pre-trained Language Models}

\author{
 Tom\'{a}s Vergara-Browne\thanks{Now at Mila and McGill University}\, \ \ 
 \'{A}lvaro Soto\\
 Pontificia Universidad Cat\'{o}lica de Chile\\
 CENIA\\
\small{{\tt  tomas.vergarabrowne@mila.quebec}}
}

\begin{document}
\maketitle
\begin{abstract}

Motivated by the surge of large language models, there has been a push to formally characterize the symbolic abilities intrinsic to the transformer architecture. A programming language, called RASP, has been proposed, which can be directly \textit{compiled} into transformer weights to implement these algorithms. However, the tasks that can be implemented in RASP are often uncommon to learn from natural unsupervised data, showing a mismatch between theoretical capabilities of the transformer architecture, and the practical learnability of these capabilities from unsupervised data. We propose \verb|tracr-injection|, a method that allows us to \textit{distill} algorithms written in RASP directly into a pre-trained language model. We showcase our method by injecting 3 different algorithms into a language model. We show how our method creates an \textit{interpretable subspace} within the model's residual stream, which can be decoded into the variables present in the code of the RASP algorithm. Additionally, we found that the proposed method can improve out-of-distribution performance compared to our baseline, indicating that indeed a more symbolic mechanism is taking place in the inner workings of the model. We release the code used to run our experiments\footnote{\href{https://github.com/tvergara/tracr-injection}{https://github.com/tvergara/tracr-injection}}.

\end{abstract}

\section{Introduction}

\begin{figure}
    \centering
    \vspace{-0.2in}\includegraphics[width=1\linewidth]{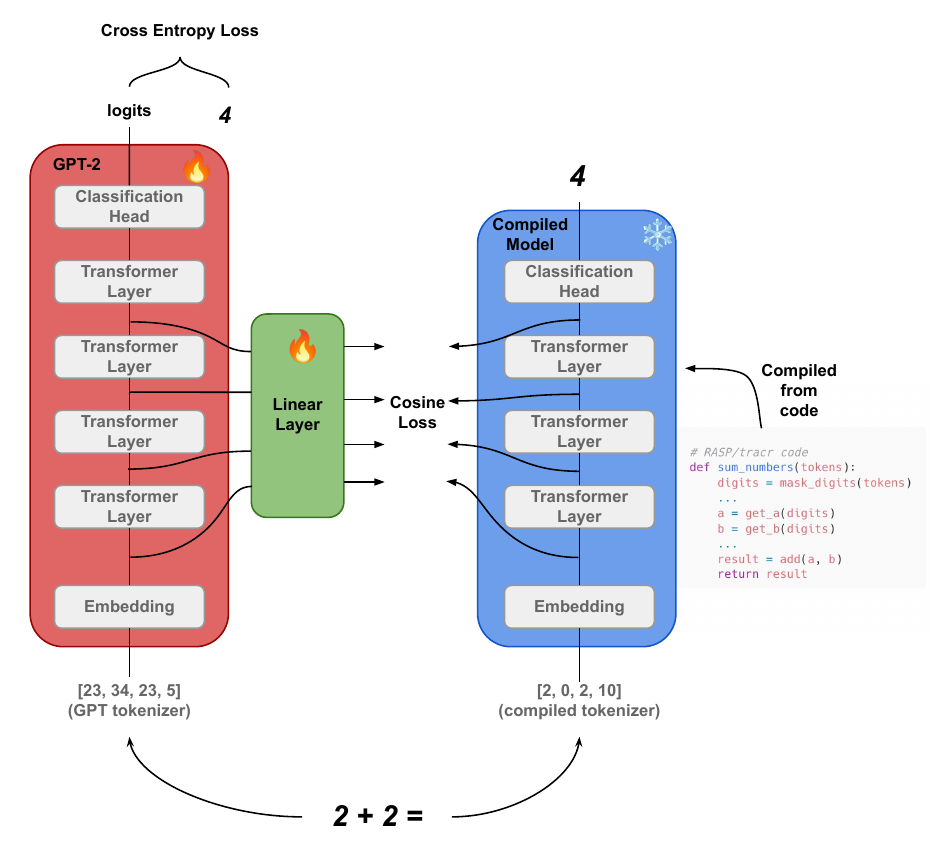}
    \caption{An overview of proposed method: we distill the residual stream of a compiled transformer using tracr, into a pre-trained model. We enforce the residual stream to \textit{linearly encode} the compiled model's representations.}
    \label{fig:injection}
\end{figure}

Transformer \cite{vaswani2017attention} based language models have become the main tool in NLP research. Recently, there has been growing interest in formally characterizing the types of mechanisms that this architecture can represent within its parameters. A key contribution in this research landscape is RASP \cite{weiss2021thinking}, a theoretical programming language whose main feature is its ability to map code instructions into weights of a transformer layers that implement the instructions. Additionally, \verb|tracr| \cite{lindner2024tracr} is a Python library that implements RASP compilation into actual transformer networks.

Many algorithms that are theoretically possible to implement in RASP (and therefore in a transformer), are not necessarily easy to learn for language models. An example is the recognition of \textit{Dyck-n languages} \cite{ebrahimi2020can}, which involves recognizing the correct balance of $n$ different parentheses within a string. For example, "([\{\}()])" correctly belongs to Dyck-3, while "([\{\}(]))" does not. Some work has explored the difficulty of learning to recognize such languages in toy transformer networks \cite{ebrahimi2020can}, and pre-trained language models tend not to do well to recognize such data without undergoing extra finetuning \cite{zhang2024darg}.

Previous work \cite{weng2023mastering} improved the symbolic capabilities of language models by offloading token generation to a compiled RASP model in symbolic tasks. However, this approach relies on handcrafted rules to route token generation, raising questions about its advantage over simpler methods like directly executing external code. In contrast, we propose to \textit{distill} algorithms implemented in RASP directly into the parameters of a pre-trained language model. Our method enhances the model's symbolic abilities \textit{internally}, reducing inference complexity and directly boosting intrinsic symbolic capabilities. Moreover, our method provides direct access to \textit{ground truth} variables underlying model predictions, significantly helping interpretability research, especially in automated circuit discovery \cite{conmy2023towards}. These interpretability techniques aim to identify causally relevant components responsible for specific computations (e.g., the components responsible for computing additions) within language models. Evaluating these methods is challenging without ground truth references \cite{mosbach2024insights}. Our method addresses this concern by explicitly injecting known algorithms as ground truth in pre-trained language models, enabling a novel way to benchmark and validate these interpretability methods. To the best of our knowledge, our work is the first to attempt to bridge this gap.

We summarize our contributions as follows:
\begin{itemize}
    \item We introduce \verb|tracr-injection|, a method to distill algorithms written in RASP into the inner workings of a pre-trained LLM.
    \item We demonstrate how to decode the residual stream \cite{elhage2021mathematical} of the final model into interpretable RASP variables, enhancing the interpretability of the mechanisms in the model.
    \item We show that the method uses the distilled representations from tracr models causally in their predictions, injecting a "ground truth" representation which the model causally uses to solve the task.
\end{itemize}

\section{Related Work}

\paragraph{RASP} RASP, introduced by \citet{weiss2021thinking}, is a theoretical programming language that maps code instructions directly to transformer weights. The main motivation of RASP was to provide a lower bound on the types of algorithms that are theoretically possible for transformer-based models to implement, and indeed some families of algorithms have been proven to be implementable inspired on this research \cite{yang2024counting, merrill2023logic, zhou2023algorithms}. Building on RASP, \verb|tracr| \cite{lindner2024tracr} provides a Python library to compile RASP algorithms into transformer weights. Recently, new alternatives to RASP and tracr have been proposed. ALTA \cite{shaw2alta} has also been proposed as an alternative programming language to RASP, with its own compiler publicly available. \citet{smolensky2024mechanisms} propose PSL, a Turing complete language which is also capable to be mapped to transformer weights.

\textbf{Distillation} Distillation is a well-established technique in machine learning \cite{hinton2015distilling}. Our approach uses a layer-wise distillation method—successfully applied in prior works \cite{romero2014fitnets, jiao2019tinybert}—to inject tracr-compiled representations into a pre-trained LLM. To our knowledge, we are the first to distill compiled models using \verb|tracr|. Perhaps the work most similar in spirit is \citet{geiger2022inducing}, in which given a causal computational graph, they enforce machine learning models to encode the variables of this graph, and also train the model to follow the same causal relationships between the variables. They also test in o.o.d. settings to verify that their training procedure enforces a stronger generalization than a naive training. Although they do not categorize their method a distillation method, it is highly related.

\textbf{Automatic Circuit Discovery} In interpretability research, a machine learning model can be seen as a computational graph \cite{geiger2021causal}, in which only a subgraph is responsible for a specific ``mechanism'' which might be human-understandable \cite{geiger2023causal}. Automated Circuit Discovery techniques \cite{conmy2023towards}, try to automatically compute the responsible subgraph for a mechanism, finding a minimal subset of nodes and edges to keep to explain the full behaviour of the model in some distribution of data. Research in this area has gained significant attention \cite{syedattribution, marks2024sparse, ferrando2024primer, bhaskar2024finding}. However, the lack of a ground truth as to which is the ``correct subgraph" is a significant limitation of this research area \cite{mosbach2024insights}. Our method is related to this area in the sense that we try to address this concern by injecting a ground truth algorithm into the weights of a pre-trained model.

\section{Methods}
\label{methods}

A diagram of our approach is shown in Figure \ref{fig:injection}. Our \verb|tracr-injection| method for injecting an algorithm into an LLM follows this procedure:
\begin{itemize}
    \item We program and compile the desired algorithm using \verb|tracr|. From this, we obtain a transformer model (and its tokenizer) which correctly implements the algorithm.
    \item We finetune a pre-trained LLM to \textit{linearly encode} the residual stream of the compiled algorithm. This is done with a linear projection in the residual stream of the pre-trained model, and a cosine loss when comparing the projection to the residual stream of the compiled model. Figure \ref{fig:injection} illustrates this process.
\end{itemize}

Specifically, let $h^i_\text{model}$ be the residual stream activations (also known as \textit{hidden states}) after the layer $i$ in the model and let $h^i_\text{mapped}$ be the same values but linearly mapped into a smaller dimension. Similarly, let $h^i_\text{compiled}$ be these activations but in the model compiled with \verb|tracr|. Let $k$ be the number of layers of the compiled model. Our final loss is a sum of three components:
\begin{itemize}
    \item \(L_{CE}\): A cross-entropy loss on task data.
    \item \(L_{ALG}\): An algorithm loss, aligning the pre-trained model’s residual stream to the compiled model, defined as \\ \( \frac{1}{k}\sum_{i=0}^k 1 - \cos(h_{\text{mapped}}^i, h_{\text{compiled}}^i)\)
    \item \(L_{KL}\): A KL-divergence loss on unsupervised text from Fineweb \cite{penedo2024fineweb}.
\end{itemize}
The necessity of the \(L_{KL}\) loss arises from trying to maintain the model as close as possible as its pre-trained version, and avoid losing the general capabilities learned in pre-training. We balance the losses with hyperparameters \(\alpha,\beta,\gamma\) (in practice we set \(\alpha=\beta=\gamma=1\)).
\begin{equation}\label{eq:loss}
  L = \alpha L_{CE} + \beta L_{ALG} + \gamma L_{KL}.
\end{equation}
To further enforce reliance on the learned symbolic subspace, we modify the final residual stream in $k$ (after the last layer where there is an effect of the \(L_{ALG}\) loss) by projecting \(x\) onto the subspace defined by the linear layer (\(h_{\text{proj}} = P_{\text{linear}} h_\text{model}^k\)) and replacing its orthogonal component with the orthogonal component of another element in the batch:
\begin{equation}\label{eq:intervention}
h' = \texttt{roll}(h_\text{model}^k - h_{\text{proj}}) + h_{\text{proj}},
\end{equation}
where \(\texttt{roll}\) shifts activations along the batch dimension (using \texttt{detach} to prevent cross-batch gradients). This encourages the model to utilize the injected symbolic representation subspace, without altering the activation distribution too much.

\section{Algorithms}
We implement in RASP algorithms for 3 tasks and compile them using \verb|tracr| (with the \verb|causal=True| option which compiles it into a causal transformer). These algorithms are:
\begin{itemize}
    \item Recognition of Shuffle-Dyck: Shuffle-Dyck is a variation of dyck languages, in which the order between different parentheses is not a restriction, but only the balance of each one. For example, the string "[(])" does belong to Shuffle-Dyck. This language has been mostly studied in the RASP literature \cite{weiss2021thinking}, for its relatively easiness of implementation in RASP.

    \item Counting characters in a text: The transformer architecture is able to implement a specific counting logic \cite{yang2024counting}, however pre-trained language models have been shown on struggling to perform counting tasks \cite{yehudai2024can}. Therefore we add as a task the counting of "x" in a sequence of characters like "x x y x y y x y".

    \item Adding two numbers: Adding two numbers (as "312 + 89 =") in a causal transformer is non-trivial, and indeed many pre-trained LLMs are known to struggle with this sort of data \cite{qian2022limitations}.
\end{itemize}

\section{Experiments}

We distill each of these algorithms into GPT-2-large \cite{radford2019language}. We compare our approach with a strong baseline. This baseline consists of ablating the $L_{ALG}$ loss from the loss function in equation \ref{eq:loss}, and removing our intervention described in equation \ref{eq:intervention}. This baseline optimizes finetuning cross-entropy loss ($L_{CE}$) and KL loss on unsupervised text ($L_{KL}$), and hence we coin the baseline as Finetuning + KL. This baseline is expected to solve the tasks successfully and outperform our method in terms of distribution performance. This is due to the fact that the baseline does not contain the alternative objective $L_{ALG}$ loss which does not directly optimize the in distribution performance. Additionally, it does not face the interventions described in Section \ref{methods} during training, which severely limit the types of mechanisms the model can potentially learn during training to successfully solve the task. However, in the baseline, there are no guarantees regarding the mechanism the model is utilizing to solve the task, so there is no way to identify causal variables encoded in the model actually used.

To measure whether our method introduces catastrophic forgetting \cite{kirkpatrick2017overcoming}, we evaluate the final models on tasks that probe both classification and language-modeling abilities:

\vspace{-0.05in}
\paragraph{Classification retention} We run inference on the binary GLUE \cite{wang2018glue} tasks SST-2 \cite{socher2013recursive} and MRPC \cite{dolan2005automatically}, selecting the class with the highest logit.
\vspace{-0.05in}

\paragraph{Next-word prediction} We measure accuracy on LAMBADA \cite{paperno2016lambada}, which tests long-range contextual understanding.
\vspace{-0.05in}

\paragraph{Language-model perplexity} We compute perplexity on Tiny-Shakespeare \cite{blog2015unreasonable}, an unsupervised corpus disjoint from the finetuning data.
\vspace{0.02in}

Only the perplexity measured on the o.o.d. Tiny-Shakespeare corpus appears in the main text (Table \ref{tab:main}); the remaining results are deferred to Appendix (Table \ref{tab:main-complete}). All of our experiments share the same hyperparameters, which are shown in the Appendix \ref{app:hyperparams}. We run each experiment with the same seed for reproducibility of our results.

\section{Results}

\begin{table}[htbp]
    \centering
    \begin{tabular}{lrr}
        \toprule
        \multicolumn{3}{c}{\textbf{Shuffle Dyck}} \\
        \cmidrule(lr){1-3}
        \textbf{Method} & \textbf{Accuracy ($\uparrow$)} & \textbf{Perplexity ($\downarrow$)} \\
        \midrule
        No editing         & 0.0\%    & 569.7  \\
        Finetuning + KL    & 100.0\%   & 569.3  \\
        Tracr Injection    & 99.9\%   & 569.9  \\
        \midrule
        \multicolumn{3}{c}{\textbf{Integer Addition}} \\
        \cmidrule(lr){1-3}
        \textbf{Method} & \textbf{Accuracy ($\uparrow$)} & \textbf{Perplexity ($\downarrow$)} \\
        \midrule
        No editing         & 0.1\%   & 569.7  \\
        Finetuning + KL    & 99.7\%  & 570.1  \\
        Tracr Injection    & 95.0\%  & 572.4  \\
        \midrule
        \multicolumn{3}{c}{\textbf{Count}} \\
        \cmidrule(lr){1-3}
        \textbf{Method} & \textbf{Accuracy ($\uparrow$)} & \textbf{Perplexity ($\downarrow$)} \\
        \midrule
        No editing         & 0.0\%  & 569.7  \\
        Finetuning + KL    & 99.8\% & 571.2  \\
        Tracr Injection    & 99.2\% & 571.4  \\
        \bottomrule
    \end{tabular}
    \caption{Performance in distilled task accuracy, and the perplexity in an o.o.d corpus of text using GPT-2 distilled by the proposed method, the baseline, and the original GPT-2 model. A full version of this table is in the Appendix, Table \ref{tab:main-complete}.}
    \label{tab:main}
\end{table}

Our main results are shown in Table \ref{tab:main}, where we only show the accuracy in the injected task, compared to the perplexity in a corpus of text not used during training \cite{blog2015unreasonable}. As we see, in all of our scenarios, the final accuracy in the test set shows a good performance, both for our baseline and our proposed method. 

Also, we highlight that performance on all unrelated tasks remains high in the baseline, as evidenced by the perplexity scores. Additionally, the performance on 3 tasks is being maintained showed in our extended results in the Table \ref{tab:main-complete} of the appendix. This serves as evidence that our method does not induce catastrophic forgetting while distilling an algorithm into the model.

\section{Analysis}
\subsection{Symbolic Representations Learned}
\begin{figure}
    \centering
    \includegraphics[width=\linewidth]{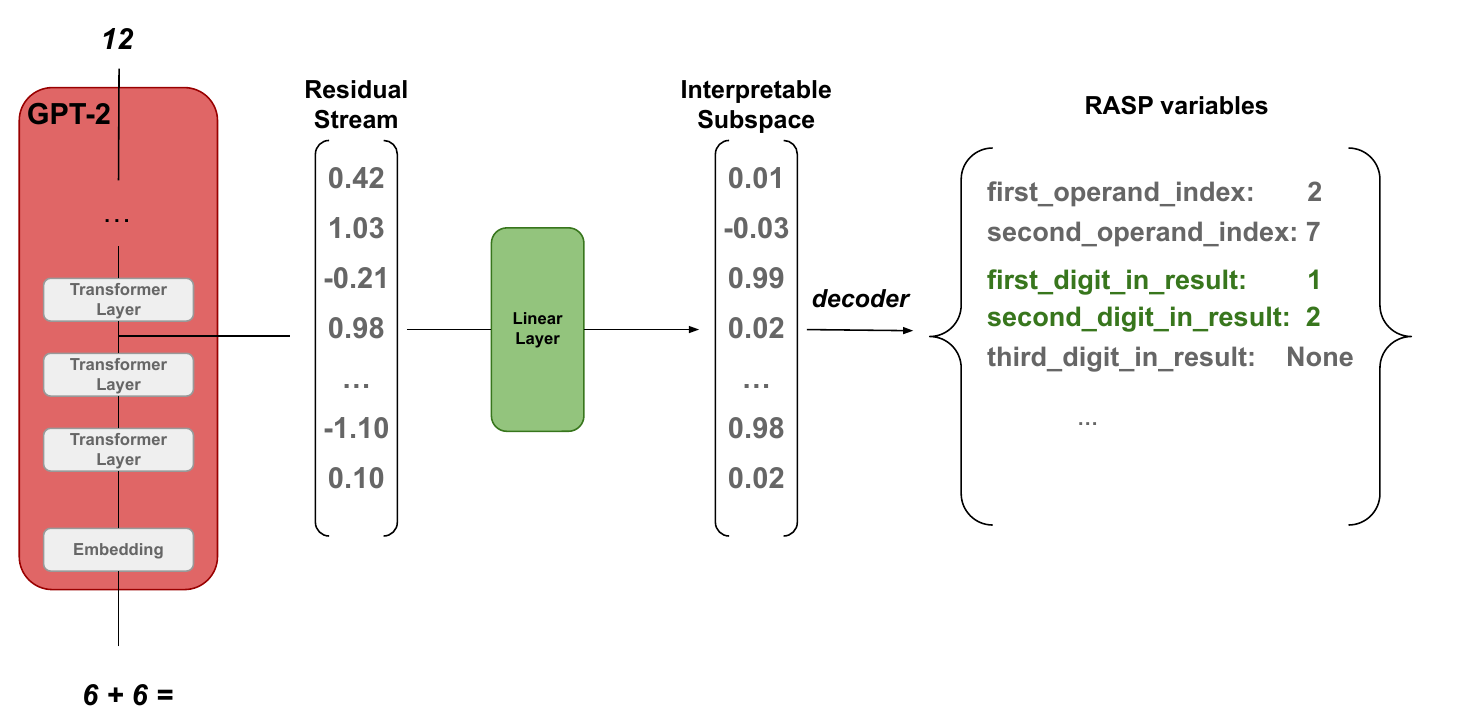}
    \caption{An example of how we can interpret the residual stream of the final model given the final linear layer used during training. Our decoder is given directly from our compiled model, where we know how each dimension in the residual stream maps to a particular value of a variable in the RASP program. An example of this process is found in Appendix \ref{app:walkthough}.}
    \label{fig:decoding}
\end{figure}

A key motivation for our work is to enforce symbolic reasoning in a pre-trained language model by injecting explicit RASP variables into its activations. We first verify that the model encodes these variables in its residual stream: our learned linear layer maps the residual stream activations to the compiled model's representation, enabling us to decode intermediate variables using their known labels (see Figure \ref{fig:decoding} and Appendix \ref{app:walkthough}).

To assess their causal role, we perturb the symbolic subspace by injecting noise. This degradation in performance is significantly larger than that caused by perturbing random directions, indicating that the injected symbolic representations are causally influential in the model's predictions (see Appendix \ref{app:causal}).

\begin{table}[htbp]
    \centering
    \small
    \begin{tabularx}{\columnwidth}{lXrrc}
    \toprule
        \textbf{Task} & \textbf{Test Data} & \textbf{TI} & \textbf{FT} \\
        \midrule
        Integer Sum  & Cascading Overflow         & 47.8\% & 49.1\% \\
        Integer Sum  & 4 digit sums               &  0.9\% &  6.6\% \\
        Integer Sum  & Decimals                   &  0.0\% &  0.0\% \\
        Shuffle-Dyck & \textit{almost-balanced} examples & 96.6\% & 75.28\% \\
        Shuffle-Dyck & 3x length                  & 98.5\% & 88.6\% \\
        Shuffle-Dyck & Additional parentheses family & 99.1\% & 84.3\% \\
        Count        & X only data                & 21.9\% & 33.3\% \\
        Count        & 3x length                  &  2.8\% &  3.5\% \\
        Count        & Replace y $\to$ z          &  4.6\% &  3.0\% \\
        \bottomrule
    \end{tabularx}
    \caption{Performance in out-of-distribution scenarios comparing Tracr-Injection (TI) and our finetuning + KL baseline (FT). A full description of each setting is in Appendix \ref{app:ood-examples}.}
    \label{tab:ood}
\end{table}
The encoding of ground truth variables that the model causally uses for its predictions is an important benefit of our proposed method. By being able to decode the residual stream activations into these interpretable variables, we are much closer to having a final model that we can understand \textit{how} it solves the task. This is not the case with normal finetuning settings, including our baseline, as we have no strong methods to extract ground-truth representations which are causally relevant to solve the task at hand. The call for more transparent systems has been raised several times in recent years \cite{bengio2025international, casper2024black}, and our method is a step toward this goal.

\vspace{-0.07in}

\subsection{Tracr-Injection can enhance o.o.d. generalization}

A final test that would let us comfortably claim that we injected a ground truth algorithm into the model would be if the final model is able to make consistent predictions outside of the data it was initially trained on, showing that it is able to implement the RASP algorithm in its whole spectrum.

We design several out-of-distribution (o.o.d.) evaluations for each algorithm. Some of these tests are tasks that the compiled RASP model would solve perfectly, while others fall outside the scope of the compiled model due to the nature of the implemented algorithm. We expect Tracr-injection to be able to generalize to some extent in o.o.d. scenarios where the compiled transformer model supported the scenario, while not being able to generalize in other cases.

We show our results in Table \ref{tab:ood}. We find that our method can enhance the o.o.d. generalization in some settings. In the case of \textit{shuffle-dyck}, the accuracy in all out-of-distribution settings practically did not suffer, while the baseline considerably achieved much lower scores. In other settings, the difference between our proposed method and the baseline is not as strong, perhaps suggesting that the overall o.o.d. ability is linked to the specific algorithm which is being distilled. We are optimistic that future work more focused on distilling the causal structure from the compiled network \cite{geiger2022inducing} can have the potential to overcome these limitations and make o.o.d. capabilities generalize independent of the algorithm at hand.

\vspace{-0.07in}

\section{Future Work}
\vspace{-0.07in}

Our method provides the first step towards truly injecting a ground truth mechanism into a pre-trained language model. However, the lack of consistent o.o.d. generalization highlights that more work needs to be done in order to truly inject a mechanism that generalizes. One avenue of exploration is understanding why certain injected algorithms seem to generalize consistently better than other algorithms (as the shuffle-dyck algorithm). There might be connections between these results and RASP-L \cite{zhou2023algorithms}. As RASP-L is a "learnable" subset of RASP instructions, it might be that some algorithms were inherently ``harder" to truly learn than others. Additionally, scaling our method to more challenging tasks is a fruitful avenue forward, as all of our tasks could be easily learned by both the baseline and our method. Scaling to these more complex tasks comes with its own complications (as the implementation of ever more complex RASP programs), but it can highlight the benefits of using Tracr-Injection in comparison to finetuning.

Additionally, the possibility to inject a ground truth mechanism into pre-trained LLMs is exciting for interpretability research. A big problem in interpretability research is the lack of "ground truth". This is a limitation when evaluating methods that, for example, automatically extract a "circuit" for a task \cite{conmy2023towards}. This lack of ``ground truth" is in fact a common criticism in interpretability research in NLP \cite{mosbach2024insights}. With the possibility of injecting a ``ground truth" to LLMs inner workings, a door is opened to start benchmarking interpretability methods based on injected ground truths.

\section{Limitations}
   
   While our methods do provide a way of distilling the RASP variables into the residual stream of a transformer, the out-of-distribution tests suggests the model is not representing these same variables when tested in examples which are radically outside of the distribution seen in training. This is a big limitation, as a fully injected algorithm should work in all distributions of inputs, and not just in a vicinity. Additionally, the use of 3 simple symbolic tasks to showcase our method is another limitation. The injection of more complex RASP programs can potentially highlight the performance benefits of our method in constrast to finetuning. The final limitation we mention is the focus exclusively on GPT-2 to evaluate the performance of our work. Compute is an important restriction for which we have made this decision, and we encourage further work in scaling up our methods into more (and larger) language models.

\section{Acknowledgements}
   
   This project was funded by the National Center for Artificial Intelligence CENIA FB210017, Basal ANID.

\bibliography{acl_latex}

\appendix

\begin{table*}[htbp]
    \centering
    \begin{tabularx}{\textwidth}{Xrrrrr}
        \toprule
        \multicolumn{6}{c}{\textbf{Shuffle Dyck}} \\
        \cmidrule(lr){1-6}
        \textbf{Method} & \textbf{Accuracy ($\uparrow$)} & \textbf{Perplexity ($\downarrow$)} & \textbf{SST-2 ($\uparrow$)} & \textbf{MRPC ($\uparrow$)} & \textbf{LAMBADA ($\uparrow$)} \\
        \midrule
        No editing         & 0.0\%    & 569.7  & 67.4\% & 63.7\% & 58.8\% \\
        Finetuning + KL    & 100.0\%   & 569.3  & 65.1\% & 58.8\% & 58.8\% \\
        Tracr Injection    & 99.9\%   & 569.9  & 75.2\% & 66.7\% & 58.8\% \\
        \midrule
        \multicolumn{6}{c}{\textbf{Integer Addition}} \\
        \cmidrule(lr){1-6}
        \textbf{Method} & \textbf{Accuracy ($\uparrow$)} & \textbf{Perplexity ($\downarrow$)} & \textbf{SST-2 ($\uparrow$)} & \textbf{MRPC ($\uparrow$)} & \textbf{LAMBADA ($\uparrow$)} \\
        \midrule
        No editing         &     0.1\%     & 569.7  & 67.4\% & 63.7\% & 58.8\% \\
        Finetuning + KL    &        99.7\%  &    570.1    &     67.2\%   &  66.2\%      & 58.7\% \\
        Tracr Injection    &       95.0\%  &    572.4    &     65.0\%   &  65.9\%      & 58.9\% \\
        \midrule
        \multicolumn{6}{c}{\textbf{Count}} \\
        \cmidrule(lr){1-6}
        \textbf{Method} & \textbf{Accuracy ($\uparrow$)} & \textbf{Perplexity ($\downarrow$)} & \textbf{SST-2 ($\uparrow$)} & \textbf{MRPC ($\uparrow$)} & \textbf{LAMBADA ($\uparrow$)} \\
        \midrule
        No editing         &      0.0\%    & 569.7  & 67.4\% & 63.7\% & 58.8\% \\
        Finetuning + KL    &     99.8\%     &    571.2    &    71.0\%    &   61.8\%     & 58.7\% \\
        Tracr Injection    &      99.2\%    &    571.4    &      64.0\%  &    65.7\%    &  58.8\%\\
        \bottomrule
    \end{tabularx}
    \caption{Performance in distilled task accuracy, and unrelated tasks using GPT-2 distilled by the proposed method, the baseline, and the original GPT-2 model.}
    \label{tab:main-complete}
\end{table*}

\section{Learned variables are causally influential}
\label{app:causal}

Although our motivation is to make the models prediction to function symbolically, and we have shown that the model effectively encode the symbolic representations that we want, these encodings might not be \textit{causally} involved in the final prediction of the model. To verify that the models use these encodings causally, we design an experiment where we add noise in the residual stream of the model at the last layer which the pre-trained model had an algorithm loss (in which it should already have all the symbolic variables encoded). We compare two ways of adding noise:
\begin{itemize}
    \item Adding noise on the direction of were the answer is encoded in the linear layer.
    \item Adding noise in random directions (but the same number of dimensions as the previous subspace).
\end{itemize}

\begin{figure*}
    \centering
    \includegraphics[width=1\linewidth]{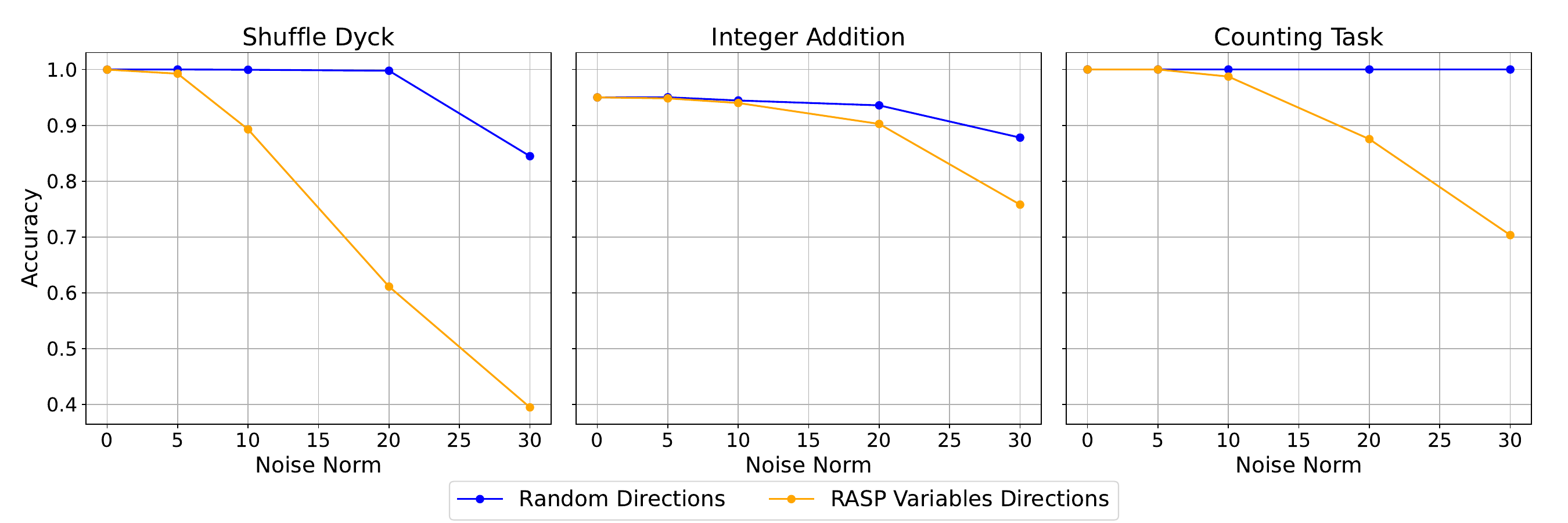}
    \caption{Accuracy in a model distilled with our algorithms. We add noise of different norms in the residual stream in the subspace where the final RASP variables are being encoded encode, and compare it to adding noise in random directions. A considerably larger effect is being made by the noise in the RASP variables subspace.}
    \label{fig:causality}
\end{figure*}

Our results for this experiment on \textit{shuffle-dyck} is shown in Figure \ref{fig:causality}. We see that for any magnitude of noise, adding noise in the directions of were the answer is encoded in the linear layer is more hurtful towards the loss, than in random directions. This suggests partially that the model is using this direction causally in its predictions, more than random directions.

\section{Hyperparameters}
\label{app:hyperparams}

For each experiment we use the exact same hyperparameter settings:
\begin{itemize}
    \item 80k different training batches of synthetic data on the task.
    \item Learning rate of $10^{-6}$ in GPT-2, and $10^{-4}$ in the linear layer.
    \item Batch size of $12$ task examples and $12$ examples of unsupervised text from \citet{penedo2024fineweb}.
    \item Number of tokens per batch example of unsupervised data of 70.
    \item The seed used was 42.
\end{itemize}
Each training run was made on a single A40 48 GB GPU for approximately 10 hours.

\section{Details on each implemented task}
\label{app:examples-task}
We explain an example of each implemented task below:
\begin{itemize}
    \item \textbf{Count task}
    \begin{itemize}
        \item Example: "What is the number of 'X' in this text? x y x x y. Answer: 3"
    \item We only do supervision on the final answer tokens (the 3 in this case). We generate a random selection of all the possibilities of cases of this task up until 40 total "x" or "y" characters, such that no example is repeated in training.
    \end{itemize}
    \item \textbf{Integer sum task}
    \begin{itemize}
        \item Example: "211 + 123 = 334"
    \item We only do supervision on the final answer tokens (the 334 in this case). We generate all the combinations of sums of values until 450, which implies that the final result is not more than 3 digits (our compiled model with RASP manages up until 3 digit sums). Again, no example is repeated during training.
    \end{itemize}
        \item \textbf{Shuffle Dyck task}
    \begin{itemize}
        \item Example: "Are parentheses here correctly matched? []\{(\}).Answer: Yes"
    \item We only do supervision on the final answer tokens (the 'yes' token in this case). This example is a correct one because in \textit{shuffle dyck} the order between different parentheses families is not a restriction (so it does not matter that we close the "\}" before closing the $)$). We generate random examples from which half are correct, and the other half are incorrect with sizes up to 30 parentheses in total. The parentheses pairs that we use are: "()", "\{\}" and "[]". Again, no example is repeated during training.
    \end{itemize}
\end{itemize}

Additionally, in number related tasks (count and integer sum), we had to make the compiled tokenizer compatible with GPT-2 into how does it separate the training data. This implied removing multidigit tokens from the tokenizer (for example, if the string "123" was getting tokenized as only one token, we remove the token from tokenizer's vocabulary such that it would get tokenized in 3 different tokens, one for each digit).

\section{Out-of-distribution examples}
\label{app:ood-examples}

Below we describe the examples we used in each of our evaluations on out-of-distribution settings. We chose 3 settings for each task, with one of these settings being a "tracr-supported" setting, which means that the original tracr-compiled model from the RASP code would solve this setting perfectly.

\begin{itemize}
    \item \textbf{Count task:}
    \begin{itemize}
        \item X-only data: we use 40 examples with only "x" values, so for example "x x x x x", from 1 to 40 "x"s (as it was trained to only see sequences up to 40). It has to count every single character. These examples were not seen in training. The compiled model can solve this scenario.
        \item 3x length: we draw examples from 41 up to 120 characters long (3x with respect to the data it was trained on). It is not supported by the tracr model as it was compiled to manage sequences up to 60 of context length.
        \item Replace 'y' with 'z': We use the original test set, but replacing the 'y' characters with 'z'. This should be an invariant in the model. The compiled model would be able to solve this, as these tokens do not affect its computation.
    \end{itemize}

     \item \textbf{Integer Sum task:}
    \begin{itemize}
        \item Cascading Overflow: We test it on data which the model has to overflow digits all the way from the right to the left, for example 299 + 1 (the model has to push the overflow digit all the way to the beginning). It is a main difficulty in implementing a sum algorithm in RASP, so we expected our method to work better. The compiled model can solve this scenario.
        \item 4 digit sums: We test it in data where each input is 4 digits long, and in training the inputs where at most 3 digits long. Our compiled model does not solve this scenario, as it was compiled to manage inputs up to 3 digits.
        \item Decimal sums: We test it in data where we append a decimal part to each value to sum. Our compiled model does not solve this scenario, as it requires an extension of the algorithm for this.
    \end{itemize}

    \item \textbf{Shuffle Dyck task:}
    \begin{itemize}
        \item Almost-balanced examples: We create correct examples, and take away the last parentheses of the example, making it unbalanced. This evaluation tests the real decision boundaries to recognize that an example is balanced or not. Our compiled model supports this scenario.
        \item 3x length: We test in data which is up to 90 parentheses long (3x with respect to the training data). Our compiled algorithm does not support this scenario, as it can handle inputs up to 90 tokens.
        \item Additional parentheses family: We add the parentheses family "<>" which is unseen in the training data. Our compiled model does not solve this scenario, as only 3 parentheses families where specified in the compilation ("()", "\{\}", "[]"). 
    \end{itemize}
\end{itemize}

\section{Walkthrough the decoding process}
\label{app:walkthough}

While a high level explanation of the decoding process can be found in Figure \ref{fig:decoding}, we leave a detailed example of this decoding process using the use case of the \textit{shuffle-dyck} task, with an input.

First, the RASP algorithm to solve this task can be found in Algorithm \ref{alg:shuffle_dyck}. For a full understanding of the algorithm, we strongly suggest reading the RASP and tracr papers \cite{weiss2021thinking, lindner2024tracr}, as the syntax and instructions specific to RASP will become clearer. However, for the purpose of our contribution, its only important to notice the \textbf{bold variables}. Each of these variables are encoded in one-hot in a subset of the neurons in the residual stream of the compiled model. For a single variable, it has a specific value according to the token position, so in order to decode a variable we also need to set fixed the position it is being decoded at. An example is the default variable \verb|indices|, which is just the positional encoding. In an arbitrary selection of tokens like $"[12, 25, 74]"$, decoding the \verb|indices| variable would give us $"[0, 1, 2]"$. Using the example of the algorithm, the variable \verb|left_start_counts| gives us the number of "starts" seen. For example the input "(())()" would give this variable the value $"[1,2,2,2,3,3]"$.

So the process of decoding the residual stream from the trained GPT-2's residual stream becomes:
\begin{itemize}
    \item Translating the residual stream activations into the compiled model's residual stream space by using the linear layer trained.
    \item For each token, and for each variable, we \verb|argmax| the neurons corresponding to them, to get a value. 
\end{itemize}

\begin{algorithm}[t]
\caption{Shuffle-Dyck RASP code}
\label{alg:shuffle_dyck}
\SetAlgoLined
\DontPrintSemicolon
\SetKwProg{Fn}{Function}{:}{}
\Fn{\texttt{shuffle\_dyck(pairs = ['()', '\{\}', '[]'])}}{
    all\_diffs $\gets$ []\;
    all\_negs $\gets$ []\;
    
    \ForEach{(left, right) in pairs}{
        starts $\gets$ SELECT(tokens, token == left)\;
        \textbf{left\_start\_counts} $\gets$ COUNT(starts)\;
        
        ends $\gets$ SELECT(tokens, token == right)\;
        \textbf{right\_end\_counts} $\gets$ COUNT(ends)\;
        
        \tcp{Compute diff between starts and ends}
        \textbf{left\_right\_diffs} $\gets$ \textbf{left\_start\_counts} $-$ \textbf{right\_end\_counts}\;
        Append \textbf{left\_right\_diffs} to all\_diffs\;
        
        \tcp{Count negative differences}
        negs\_selector $\gets$ SELECT(\textbf{left\_right\_diffs}, value $<$ 0)\;
        \textbf{left\_right\_negative\_counters} $\gets$ COUNT(negs\_selector)\;
        Append \textbf{left\_right\_negative\_counters} to all\_negs\;
    }
    
    \tcp{Aggregate all negative counters}
    current\_negs $\gets$ all\_negs[0]\;
    \ForEach{negs in all\_negs[1:]}{
        current\_negs $\gets$ MAP( (x,y) $\rightarrow$ if (x $\neq$ 0 or y $\neq$ 0) then 1 else 0, current\_negs, negs )\;
    }
    \textbf{aggregated\_negatives} $\gets$ current\_negs\;
    
    \tcp{Aggregate all balances}
    current\_diffs $\gets$ all\_diffs[0]\;
    \ForEach{diffs in all\_diffs[1:]}{
        current\_diffs $\gets$ MAP( (x,y) $\rightarrow$ if (x $\neq$ 0 or y $\neq$ 0) then 1 else 0, current\_diffs, diffs )\;
    }
    \textbf{aggregated\_diffs} $\gets$ current\_diffs\;
    
    \textbf{unfiltered\_result} $\gets$ MAP( (x,y) $\rightarrow$ (x and y == 0), aggregated\_negatives, aggregated\_diffs )\;
    \textbf{final\_result} $\gets$ MAP( (x, token) $\rightarrow$ if (token == "compute") then x else False, unfiltered\_result, tokens )\;
    
    \Return \textbf{final\_result}\;
}
\end{algorithm}

\begin{figure}
    \centering
    \includegraphics[width=1\linewidth]{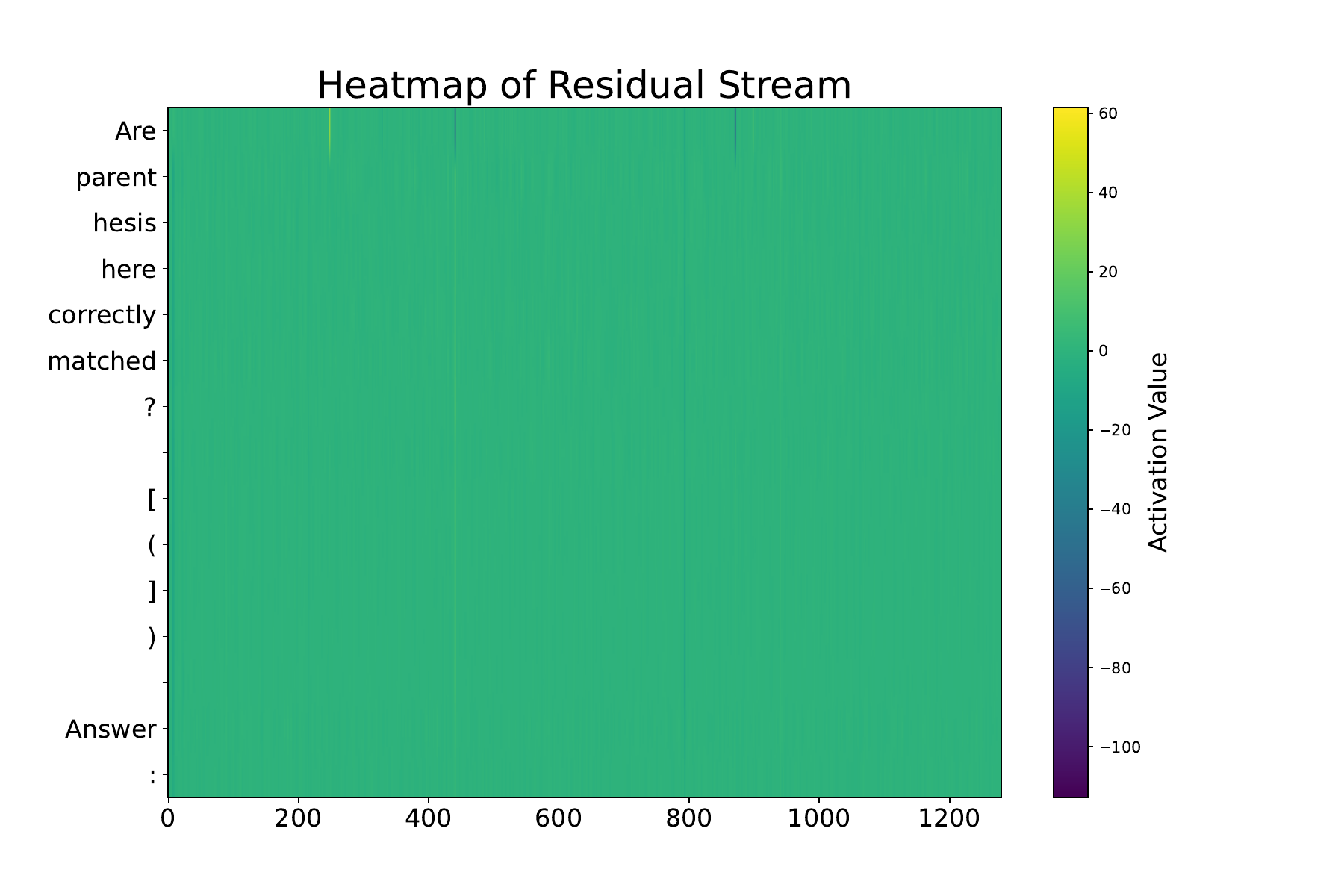}
    \caption{A plot of the residual stream in every token of the string "Are parentheses here correctly matched? [(]). Answer:" in a GPT-2 with the \textit{shuffle-dyck} distiled on it. Interpreting these activations is impossible to the naked eye.}
    \label{fig:base-res-stream}
\end{figure}

\begin{figure}
    \centering
    \includegraphics[width=1\linewidth]{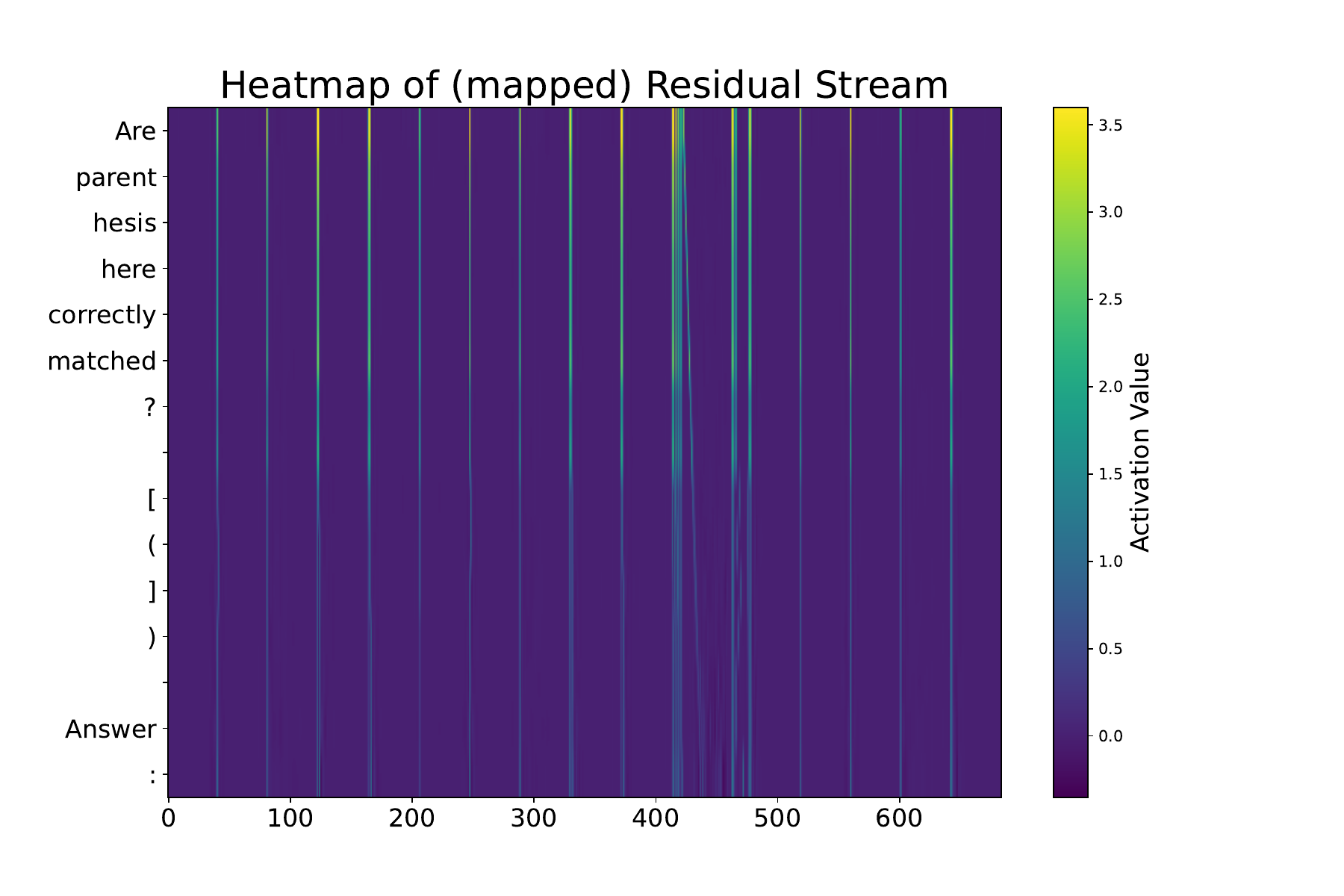}
    \caption{A plot of the residual stream in the token of the string "Are parentheses here correctly matched? [(]). Answer:", translated to a sparser space with the linear layer learned during distillation. Each straight line represents a value that is constant in every token, while other variables (like the inclined line) represent variables which change value for each token dimension.}
    \label{fig:translated-res-stream}
\end{figure}

Let us take the example of how the a GPT-2 distilled with this algorithm can correctly predict the next token in a string like: "Are parentheses here correctly matched? [(]). Answer:". A heatmap of the activations of the residual stream of the model is seen in Figure \ref{fig:base-res-stream}. We see that the activations in each token are completely uninterpretable, however, when we pass these activations through the linear layer learned during distillation, we get Figure \ref{fig:translated-res-stream}. These activations are much more sparse, and we can even see clear phenomena, as straight lines. Straight lines are values which are constant over all the tokens (many variables are indeed shared independent of the token position).

\begin{figure*}
    \centering
    \includegraphics[width=\linewidth]{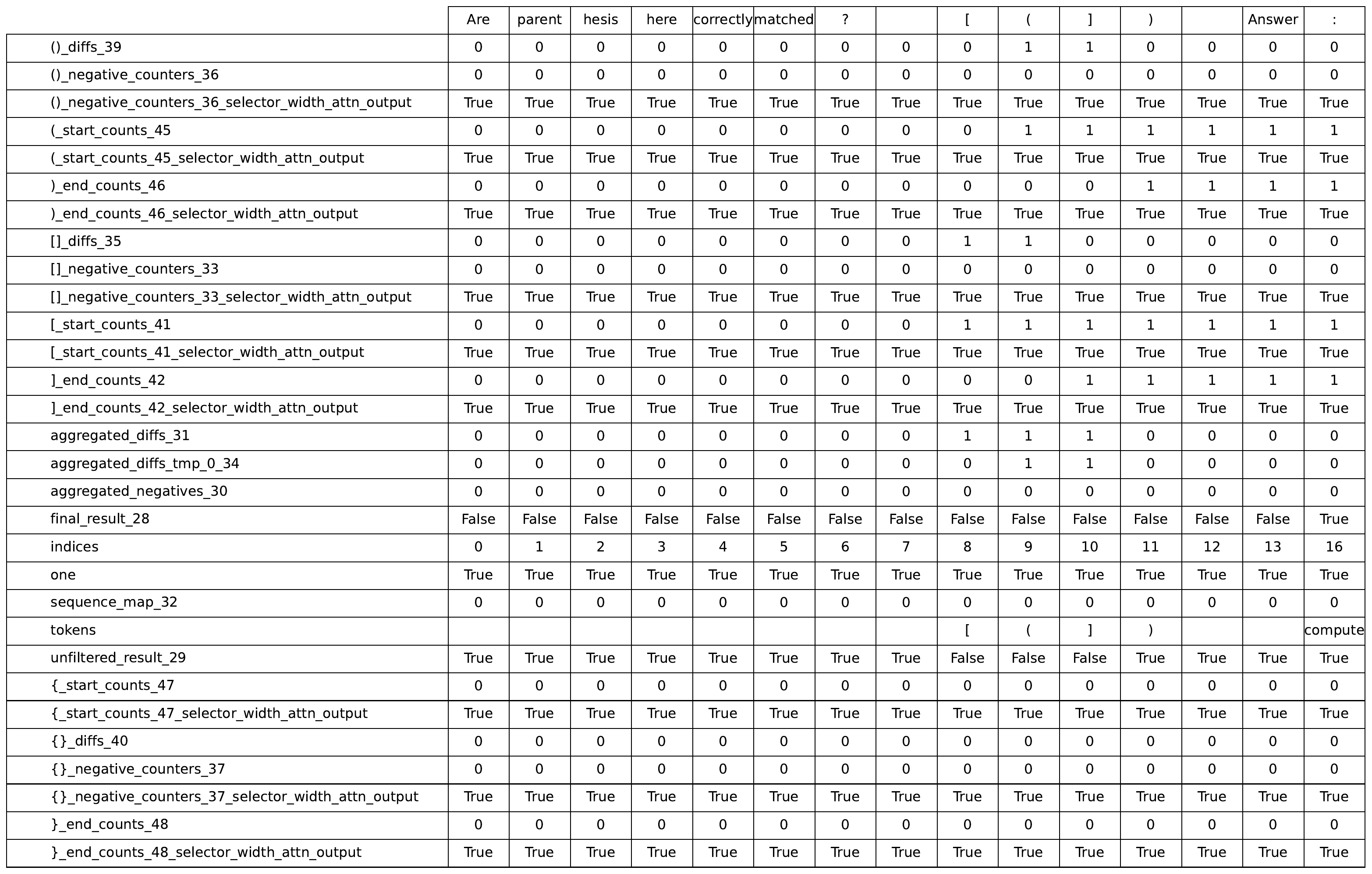}
    \caption{Decoded variables in the residual stream of a GPT-2 distilled with the Algorithm \ref{alg:shuffle_dyck} when passed through the input "Are parentheses here correctly matched? [(]). Answer:".}
    \label{fig:table-res-stream}
\end{figure*}

Finally we can map these activations into a table of the shape \verb|(tokens, variables)| as in Figure \ref{fig:table-res-stream}, where we decode the information written for each variable in each token position. This process just implies taking the argmax over the neurons encoding all the values in a variable in the residual stream, for each of the tokens.

\section{Prompts used for classification evaluation}
\label{app:prompts}
\begin{codelisting}{SST2 Prompt}
What is the sentiment in this review? The answer should be one word: either positive or negative.

Review: the movie was absolutely incredible.

Answer: positive

What is the sentiment in this review? The answer should be one word: either positive or negative.

Review: this was a shameful waste of time.

Answer: negative

What is the sentiment in this review? The answer should be one word: either positive or negative.

Review: {sentence}

Answer:
\end{codelisting}

\begin{codelisting}{MRPC Prompt}
Determine whether the following two sentences are paraphrases (i.e., have the same meaning). Answer in one word: yes or no.

Sentence 1: The company reported strong earnings this quarter.
Sentence 2: The business announced impressive profits for the quarter.
Answer: yes

Sentence 1: I love to read science fiction novels.
Sentence 2: I prefer watching documentaries over reading books.
Answer: no

Sentence 1: {sentence1}
Sentence 2: {sentence2}
Answer:
\end{codelisting}

\end{document}